\documentclass[conference]{IEEEtran}
\IEEEoverridecommandlockouts

\usepackage[dvipsnames, svgnames, x11names]{xcolor}

\usepackage{soul}
\usepackage{color}

\usepackage{cite}

\usepackage{authblk}

\usepackage{amsmath,amssymb,amsfonts}
\usepackage{algorithmic}
\usepackage{graphicx}
\usepackage{textcomp}

\usepackage{xspace}

\DeclareGraphicsExtensions{.pdf,.jpg,.png}

\usepackage{silence}
\usepackage{array}
\usepackage{booktabs} 
\usepackage{colortbl}
\usepackage{diagbox}
\usepackage{multirow}
\usepackage{tabu}
\usepackage{graphicx}
\usepackage{makecell}

\def\BibTeX{{\rm B\kern-.05em{\sc i\kern-.025em b}\kern-.08em
    T\kern-.1667em\lower.7ex\hbox{E}\kern-.125emX}}

\begin{document}

\title{FedTADBench: Federated Time-series \\Anomaly Detection Benchmark\\
\thanks{The first two authors contribute equally. Corresponding authors are Yingjie Zhou(yjzhou09@gmail.com) and Le Zhang(lezhang@uestc.edu.cn).}
}

\author[1]{Fanxing Liu}
\author[2]{Cheng Zeng}
\author[2]{Le Zhang}
\author[1]{Yingjie Zhou}
\author[1]{Qing Mu}
\author[4]{Yanru Zhang}
\author[3,5]{Ling Zhang}
\author[2]{Ce Zhu}

\affil[1]{College of Computer Science, Sichuan University, China}
\affil[2]{School of Information and Communication Engineering,\authorcr University of Electronic Science and Technology of China, China}
\affil[3]{College of Polymer Science and Engineering, Sichuan University, China}
\affil[4]{School of Computer Science and Engineering, University of Electronic Science and Technology of China, China}
\affil[5]{West China School of Public Health, Sichuan University, China}

\maketitle

\begin{abstract}

Time series anomaly detection strives to uncover potential abnormal behaviors and patterns from temporal data, and has fundamental significance in diverse application scenarios. Constructing an effective detection model usually requires adequate training data stored in a centralized manner, however, this requirement sometimes could not be satisfied in realistic scenarios. As a prevailing approach to address the above problem, federated learning has demonstrated its power to cooperate with the distributed data available while protecting the privacy of data providers. However, it is still unclear that how existing time series anomaly detection algorithms perform with decentralized data storage and privacy protection through federated learning. To study this, we conduct a federated time series anomaly detection benchmark, named FedTADBench, which involves five representative time series anomaly detection algorithms and four popular federated learning methods. We would like to answer the following questions: (1)How is the performance of time series anomaly detection algorithms when meeting federated learning? (2) Which federated learning method is the most appropriate one for time series anomaly detection? (3) How do federated time series anomaly detection approaches perform on different partitions of data in clients? Numbers of results as well as corresponding analysis are provided from extensive experiments with various settings. The source code of our benchmark is publicly available at https://github.com/fanxingliu2020/FedTADBench.

\end{abstract}


\section{Introduction}

Time series anomaly detection is a vital and fundamental task in data mining, and has been broadly employed in a variety of application scenarios, such as intelligent manufacturing \cite{weng2016selflearning}, critical infrastructure monitoring \cite{xu2018unsupervised, chen2021learning}, seismic analysis\cite{qian2022multidimensional, qian2022ground}, financial fraud detection \cite{weng2019Online} and health care \cite{zhou2019BeatGAN}. During the last decades, numerous anomaly detection models have been proposed for handling temporal data. Most of these models need to train with temporal data stored in a centralized manner. However, this condition sometimes may not be satisfied in practice due to the huge communication cost of uploading distributed data to a central server and private protection of sensitive information among individual data resource. Thus, conducting time series anomaly detection with decentralized data storage and private protection is a significant and imperative problem.

\begin{figure*}[htbp]
\centerline{\includegraphics[width=0.9\linewidth]{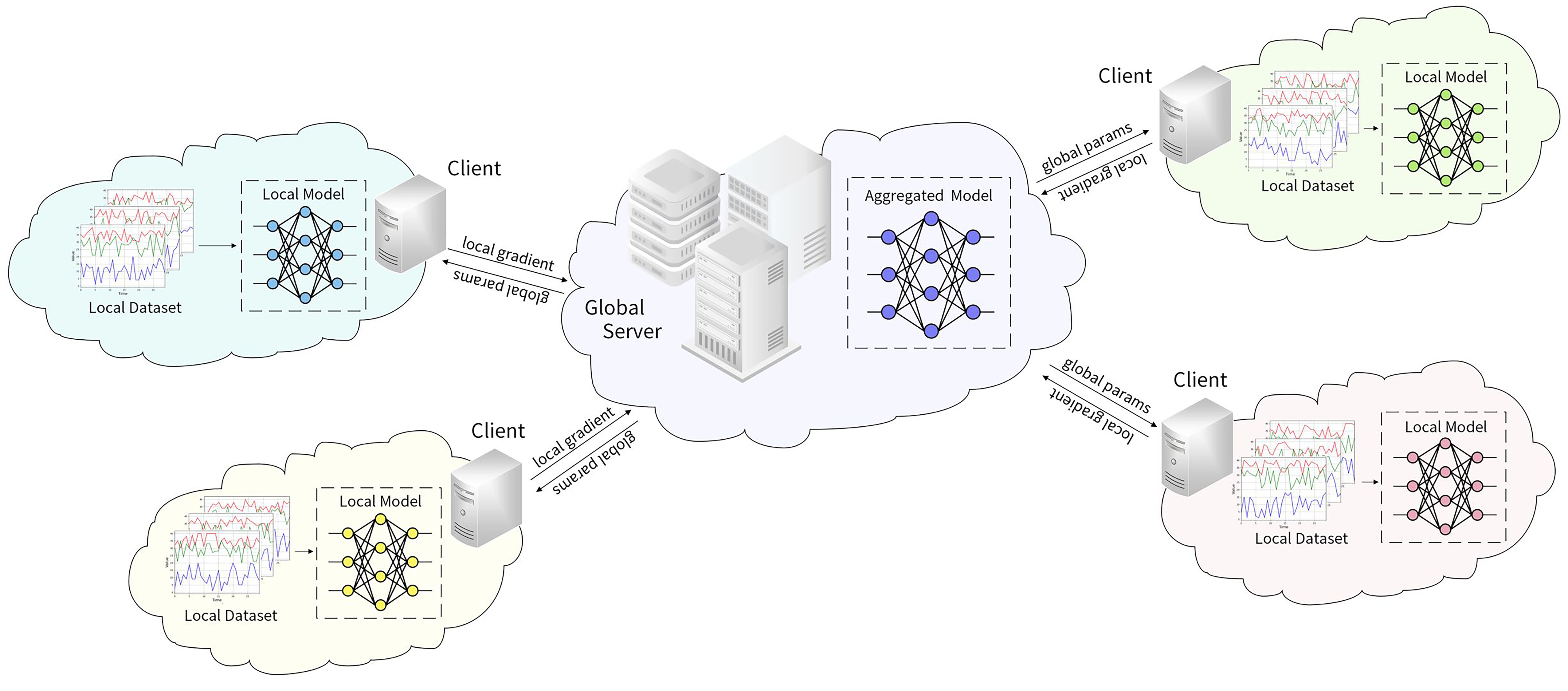}}
\caption{Overview of time series anomaly detection under the framework of federated learning.  Each client maintains its local model and trains it with its private dataset for several local epochs. Then, the local models will be sent to a centralized(global) server for calculating the updates of the global(aggregated) model. This process will be repeated for several ``global epochs" until convergence.}
\label{fig}
\end{figure*}
Federated learning\cite{mcmahan2017communication} is a learning paradigm that can conduct model training without direct access to raw data, but sharing model parameters/gradient by training clients’ data locally while protecting the privacy of clients. As a prevailing approach to address the aforementioned problem, federated learning has been widely used in various fields, such as IoT \cite{survey_iot_fed} and medical information \cite{survey_medical}, to aggregate siloed data. Some recent works have explored time series anomaly detection under the framework of federated learning(as shown in Fig. 1), such as \cite{liu2020communication}, \cite{zhang2021federated} and \cite{truong2022light}, where the feasibility of using the federated learning framework for time series anomaly detection is preliminary demonstrated. However, existing efforts mainly consider solving the specific challenges of their concerned tasks. There still lack of unbiased and in-depth evaluations of time series anomaly detection with federated learning, which could provide guideline information to construct effective models for both researchers and engineers.

To fill this blank, we construct a federated time series anomaly detection benchmark, named FedTADBench, which designs numerous evaluations for the performance of time series anomaly detection with federated learning. Specifically, we conduct experiments with three major aspects in the benchmark: (1) the feasibility of federated learning for typical time series anomaly detection algorithms, i.e., performance comparisons of time series anomaly detection algorithms with/without federated learning, and time series anomaly detection performance under isolated and federated settings; (2) the compatibility between popular federated learning methods and typical time series anomaly detection algorithms, i.e., performance comparisons of different federated learning methods for time series anomaly detection, and training time consumption of typical time series anomaly detection algorithms with different federated learning strategies; (3) the influence of the heterogeneity between clients to federated time series anomaly detection, i.e., performance comparisons of federated time series anomaly detection approaches with different partitions of data in clients. Detailed analysis is made based on the extensive experiment results.

The contributions of this paper are as follows:

\begin{enumerate}
\item We construct FedTADBench, the first federated time series anomaly detection benchmark as far as we know, which evaluates time series anomaly detection algorithms with various federated learning settings.

\item Based on evaluations from three major angles, we show valuable insights for time series anomaly detection under the framework of federated learning, which could offer guiding information for building effective models of interest.

\item We provide reproducible evaluations whose codes are fully open-source. The website is https://  github.com/fanxingliu2020/FedTADBench. 

\end{enumerate}

The rest of the paper is organized as follows. Section II introduces the related works on time series anomaly detection and federated learning, as well as the existing benchmarks. The problem statement and details of our benchmark are presented in Section III. Various experiment results and corresponding analysis are conducted in Section IV. Section V summaries the paper.

\addtolength{\topmargin}{0.05in}

\section{Related Works}

\subsection{Time Series Anomaly Detection}

The essential challenge to time series anomaly detection is formulating the pattern of temporal data characteristically, especially under certain limitations such as dearth of high-quality labeled data, heterogeneous structure of data, etc. As a result, most of existing anomaly detection models are unsupervised ones. SISVAE \cite{li2020anomaly} constructs a recurrent neural network based variational auto-encoder to model normal data, which outperforms in capturing latent temporal structures of time series. For lessening the impact of fitting rare anomalous samples, T. Kiew et al. \cite{kieu2019outlier} construct an ensemble of a series of recurrent auto-encoders with various skip connections. OmniAnomaly proposed by Y. Su et al. \cite{su2019robust} performs time series modeling by combining a stochastic recurrent neural network and a planar normalizing flow, and uses the reconstruction probabilities to determine anomalies. This work transcends most of prior methods at the cost of high training time. More recently, aiming to address the large expense of energy consumption, J. Audibert et al. \cite{audibert2020usad} propose USAD, which is capable of distinguishing anomalies from normal data rapidly with an unsupervised method by combining adversarial thought with auto-encoder.

Recently, with the increasing popularity of transformers and graph neural networks, some scholars have explored the use of them for time series anomaly detection \cite{kipf2016semi},\cite{vaswani2017attention}. Graph neural networks are used to extract correlations between dimensions in multivariate time series, while Transformers are exploited for modeling latent representations in sequences. GDN introduced by A. Deng et al. \cite{deng2021graph} combines graph neural networks by attention mechanism, and provides rationales that the detected points are determined as abnormal. TranAD proposed by S. Tuli et al. \cite{tuli2022tranad} is an unsupervised time series anomaly detection model, which is able to compute in parallel by utilizing Transformers, and gain stable results benefiting from adversarial training. Anomaly Transformer proposed by J. Xu et al.\cite{xu2021anomaly} learns global relationships of time series with attention mechanism, and utilizes the difference between local relationships and global relationships to conduct anomaly detection.

Most of the proposed approaches are under the assumption that all data is accessible to the central server for training. However, it is not always feasible in real-world scenarios where the availability of data is restricted by security and privacy concerns and there are also communication resource constraints.

\subsection{Federated Learning}
Federated learning can be categorized into vertical federated learning and horizontal federated learning \cite{fed_concept}. Here, we mainly discuss horizontal federated learning as our data share the feature space but is held by different institutions. Currently there are four classic horizontal federated learning methods: FedAvg \cite{mcmahan2017communication}, FedProx \cite{fedprox}, SCAFFOLD \cite{scaffold}, and MOON\cite{moon}. FedAvg is the first federated learning algorithm, and the others add regularization to FedAvg from different perspectives. Since federated learning has advantages in privacy protection and data communication cost, 
some researchers have conducted time series anomaly detection under the framework of federated learning. Mothukuri et al. \cite{mothukuri2021federated} adopt federated learning to construct an anomaly detection system for IoT security attacks, which trains local models on edge devices and aggregates the information on every edge device by federated learning. Liu et al. \cite{liu2020deep} proposed an attention based convolutional neural network with long short-term memory to detect edge device failure. To protect users' privacy, they adopted a federated learning framework to carry out the learning process. Besides, for improving communication efficiency, they employed a top-k selection algorithm to compress the gradient. For industrial cyber–physical systems (CPSs), it is difficult to collect sufficient high-quality attack examples. To aggregate data from different users in a privacy-preserving way, Li et al \cite{li2020deepfed} designed a federated learning system based on paillier cryptosystem to train the deep learning based intrusion detection model for industrial CPSs.

However, the previously mentioned approaches mainly focus on improving the detection performance for specific tasks. The feasibility and limitations of federated learning for time series anomaly detection tasks is still unclear.

\subsection{Existing Benchmarks}

Researchers have made sustained efforts to provide unbiased evaluations for existing anomaly detection algorithms, such as \cite{campos2016evaluation},\cite{domingues2018comparative},\cite {ruff2021unifying}, \cite{acsintoae2022ubnormal} and \cite{han2022adbench}. There have also been some reports to conduct performance comparisons of time series anomaly detection algorithms \cite{lai2021revisiting,paparrizos2022tsb}, since temporal data is commonly seen in various application scenarios. However, any benchmark for federated time series anomaly detection has not been investigated yet. This paper tries to evaluate that how do existing time series anomaly detection algorithms perform on various federated learning settings. To the best of our knowledge, this is the first benchmark for federated time series anomaly detection.

\begin{figure}[htbp]
\centerline{\includegraphics[width=0.9\linewidth]{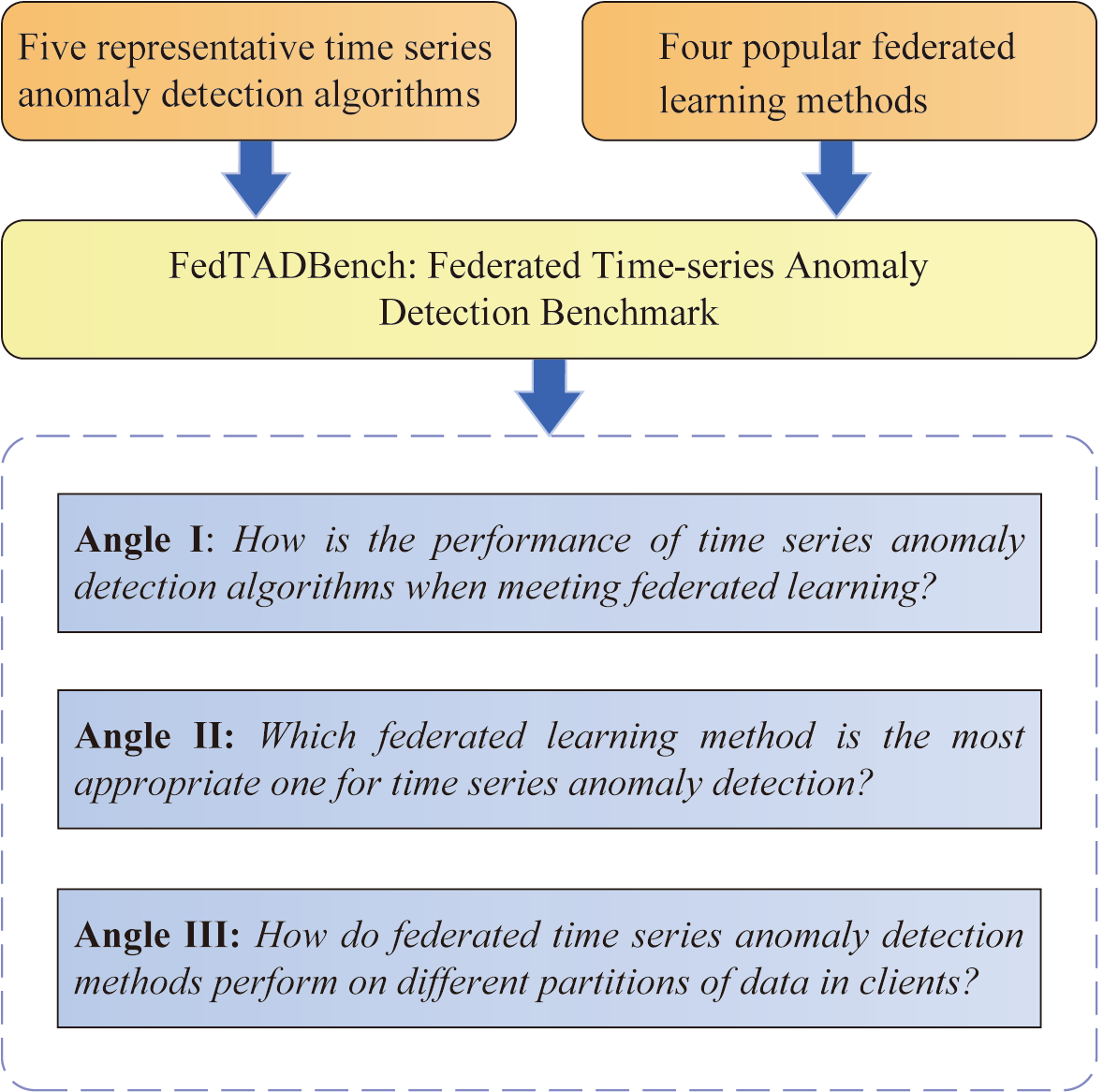}}
\caption{An overview of FedTADBench that evaluates combinations of 5 time series anomaly detection and 4 federated learning methods. It sheds light upon time series anomaly detection in federated settings from 3 angles.}
\label{fig}
\end{figure}

\section{FedTADBench: An empirical Evaluation on Federated Time-series Anomaly Detection}

We present the problem statement and elements of FedTADBench in this section. For better understanding, we provide an overview of our benchmark, as shown in Fig.~\ref{fig}, which sheds light upon time series anomaly detection in federated settings from 3 angles.

\subsection{Problem Statement}
In this study, we focus on detecting anomalies at entity-level~\cite{huang2022semi} using multivariate time series. More specifically, we consider multi-dimensional time series data which consists of multiple observations continuously collected at equal-space timestamp. The multivariate time series can be defined as $\mathbf{X}=\left[\mathbf{X}_1, \mathbf{X}_2, \cdots, \mathbf{X}_{\mathbf{n}}\right]$, where $n$ is the number of collected data, and $\mathbf{X}^{\mathrm{i}}=\left[x^i_1, x^i_2, \cdots, x^i_t\right]$ is an observation vector
of the $i^{th}$ metric within a dimensionality of $t$.  For entity-level multivariate time series anomaly detection,  our target  is to determine whether the
observation $\mathrm{X}_{\mathrm{t}}=\left[x^1_t, x^2_t, \cdots, x^n_t\right]$ at time $ t $ is anomalous or normal.

Conventional approaches learn such a function in a \emph{Centralized} manner. In this way, they usually assume  the learning algorithms have access to all the training samples. Contrastively, in many privacy-preserving scenarios, the data are usually stored in an isolated manner (namely the data can only be accessed by its owner) by different clients. And those clients are assumed to train a learning algorithm in a collaborative manner without data exchange. More specifically, in this case we assume that the data are owned by $C$ clients, i.e.,  $\mathbf{X} = \{\mathbf{X}{(1)}, \mathbf{X}{(2)},  ..., \mathbf{X}{(C)}\}$,  and the task is to collaboratively train a model in a condition that each client only can access its own data.   In a typical federated learning framework, each client of federated learning trains its \emph{local model}  with its  own private dataset for several \emph{local epochs}, then, the clients send their local models to the server. Finally,  the server aggregates the local models into a \emph{global model}, and this "training-aggregating"  regime will be repeated for several ``global epochs" until convergence.

\subsection{Elements of FedTADBench}

We evaluate 4 commonly-used federated learning methods and  5 time series anomaly detection algorithms on 3 popular time series anomaly detection datasets. In addition, we also investigate the effect of different partitions of data in clients. Below we elaborate the details.

\subsubsection{Time Series Anomaly Detection Methods}
\begin{itemize}
    \item DeepSVDD~\cite{ruff2018deep}: It jointly trains a deep neural network and  learns a data-enclosing hypersphere in the latent space for anomaly detection tasks.
    \item LSTM-AE~\cite{garg2021evaluation}. It is a combination of LSTM and Auto-encoders that modeling time series by reconstructing the original data.
    \item USAD~\cite{audibert2020usad}. It is an adversarially-trained autoencoders.  USAD is fast to train and robust to the choice of parameters. 
    \item GDN~\cite{deng2021graph}. It trains a graph neural network in a structural learning  manner. Attention weights are used as well to provide certain level of explainability for the detected anomalies.
    \item TranAD~\cite{tuli2022tranad}. It is a deep transformer network and uses attention-based sequence encoders to enable robust multi-modal feature extraction.
\end{itemize}

\subsubsection{Federated Learning Methods}

\begin{itemize}
    \item FedAvg~\cite{mcmahan2017communication}. It averages the models learned by clients to obtain the global model.
    \item FedProx~\cite{fedprox}. It adds a proximal term to regularize the parameters of the local model. In this case, the parameters of the current local model are not far from those of the previous global model.    
    \item SCAFFOLD~\cite{scaffold}.  It uses variance reduction techniques to remedy the ``client-drift" in its local updates. SCAFFOLD is shown to be more communicational efficient than FedAvg. 

    \item MOON~\cite{moon}. It regularizes the federated learning progress in a contrastive manner. More specifically, it pushes the models away from its previous status and pulls the models towards the global model.
\end{itemize}

\subsubsection{Datasets}

\begin{itemize}    
    \item SMD \cite{su2019robust}.  SMD (Server Machine Dataset) is a  5-week-long dataset and is splitted into training and testing set with  equal size. 
    \item SMAP \cite{hundman2018detecting}. 
    SMAP (Soil Moisture Active Passive satellite) is a public dataset from NASA.
    \item PSM \cite{abdulaal2021practical}. It is a dataset proposed by eBay and consists of 26 dimensional time series data from application servers.
\end{itemize}

The details of datasets that we perform experiments on are shown in Table \ref{dataset_details}. NS, ND and NC refer to the quantity of time series in each dataset,  dimensions in the time series of each dataset, and   clients in federated learning settings, respectively.

\subsubsection{Benchmark Angles}

Our benchmark is motivated by the following aspects. First, the compatibility between popular federated learning and time series anomaly detection methods is not extensively studied. Due to permutation invariance of neural network parameters\cite{fedma}, client drift in federated learning\cite{scaffold} and so on, federated learning may not be compatible with certain time series anomaly detection algorithms. So our benchmark could provide such a testbed. Second, the \emph{No-Free-Lunch}~\cite{luke2013essentials} Theorem states that within certain constraints, over the space of all possible problems, every optimization technique will perform as well as every other one on average. However, for federated learning based time series anomaly detection, this has not been explored empirically. Our benchmark can help us to verify this hypothesis as well. Third, in practice, the amount of data held by clients may vary significantly. This essentially brings in heterogeneity between the clients and usually degrades the performance. To study how the heterogeneity between clients affects the performance of time series anomaly detection, we test the effect of different levels of client heterogeneity on performance.

\begin{table}[htbp]
\caption{Dataset Details and corresponding federated learning settings.  NS, ND and NC refer to the quantity of time series in each dataset,  dimensions of the time series in each dataset, and   clients in federated learning settings, respectively.}
\begin{center}
\begin{tabular}{cccc}
\hline \qquad Dataset \qquad & \qquad NS \qquad & \qquad ND \qquad & \qquad NC \qquad \qquad \\
\hline \qquad SMD \qquad & \qquad 28 \qquad & \qquad 38 \qquad & \qquad 28 \qquad \qquad \\
\qquad SMAP \qquad & \qquad 54 \qquad & \qquad 25 \qquad & \qquad 54 \qquad \qquad\\
\qquad PSM \qquad & \qquad 1 \qquad & \qquad 25 \qquad & \qquad 24 \qquad \qquad \\
\hline
\end{tabular}
\end{center}
\label{dataset_details}
\end{table}

\begin{table*}[ht]
\caption{Detailed Results of FedTADBench in terms of AUC-ROC and AUC-PR.  We highlight the best result and the second best result  in blue and light blue, respectively.}
\begin{center}
\newcommand\mtc[1]{\multicolumn{2}{c}{#1}} 
\newcommand\mtrtotwo[1]{\multirow{2}*{#1}} 
\newcommand\linone{\Xcline{3-12}{0.4pt}}
\resizebox{\linewidth}{!}{
\begin{tabular}{cl|cc cc cc cc cc}
\Xhline{1pt}
\multicolumn{2}{c|}{\mtrtotwo{{Method}}}&\mtc{{DeepSVDD}}&\mtc{{LSTM-AE}}&\mtc{{USAD}}&\mtc{{GDN}}&\mtc{{TranAD}}\\
\makebox[10pt][c]{}&\makebox[0.03\textwidth][c|]{}&{AUC-ROC}&{AUC-PR}&{AUC-ROC}&{AUC-PR}&{AUC-ROC}&{AUC-PR}&{AUC-ROC}&{AUC-PR}&{AUC-ROC}&{AUC-PR}\\

\hline
\multirow{5}*{\rotatebox{90}{SMD}}&Original&0.6311&0.0818&\sethlcolor{LightSkyBlue}\hl{0.6231}&\sethlcolor{LightSkyBlue}\hl{0.0871}&\sethlcolor{LightSkyBlue}\hl{0.6874}&\sethlcolor{LightBlue1}\hl{0.1182}&\sethlcolor{LightSkyBlue}\hl{0.6664}&\sethlcolor{LightSkyBlue}\hl{0.0946}&0.6028&0.1120\\
&FedAvg&0.6368&\sethlcolor{LightBlue1}\hl{0.1291}&0.5915&0.0764&0.6518&\sethlcolor{LightSkyBlue}\hl{0.1197}&\sethlcolor{LightBlue1}\hl{0.6509}&\sethlcolor{LightBlue1}\hl{0.0785}&\sethlcolor{LightBlue1}\hl{0.6335}&\sethlcolor{LightSkyBlue}\hl{0.1311}\\
&FedProx&\sethlcolor{LightBlue1}\hl{0.6401}&0.0858&0.5921&\sethlcolor{LightBlue1}\hl{0.0767}&0.5451&0.0542&0.6383&0.0747&0.5720&0.0582\\
&Scaffold&0.5773&0.0909&\sethlcolor{LightBlue1}\hl{0.6185}&0.0756&\sethlcolor{LightBlue1}\hl{0.6625}&0.1096&0.6302&0.0692&0.6211&0.1087\\
&Moon&\sethlcolor{LightSkyBlue}\hl{0.6679}&\sethlcolor{LightSkyBlue}\hl{0.1612}&0.5981&0.0748&0.6433&0.0951&0.6273&0.0777&\sethlcolor{LightSkyBlue}\hl{0.6337}&\sethlcolor{LightBlue1}\hl{0.1310}\\

\hline
\multirow{5}*{\rotatebox{90}{SMAP}}&Original&0.6032&0.1549&0.4442&0.1155&0.5520&0.1312&\sethlcolor{LightBlue1}\hl{0.5369}&\sethlcolor{LightBlue1}\hl{0.1352}&\sethlcolor{LightSkyBlue}\hl{0.5754}&\sethlcolor{LightSkyBlue}\hl{0.1423}\\
&FedAvg&\sethlcolor{LightSkyBlue}\hl{0.6333}&\sethlcolor{LightBlue1}\hl{0.1841}&0.4328&0.1141&\sethlcolor{LightBlue1}\hl{0.5786}&\sethlcolor{LightBlue1}\hl{0.1418}&0.5159&0.1228&0.5392&0.1274\\
&FedProx&0.4299&0.1094&\sethlcolor{LightBlue1}\hl{0.4523}&\sethlcolor{LightBlue1}\hl{0.1168}&0.4579&0.1088&0.5218&0.1247&0.5361&0.1272\\
&Scaffold&0.5992&0.1529&0.4328&0.1141&\sethlcolor{LightSkyBlue}\hl{0.5908}&\sethlcolor{LightSkyBlue}\hl{0.1451}&\sethlcolor{LightSkyBlue}\hl{0.5761}&\sethlcolor{LightSkyBlue}\hl{0.1361}&\sethlcolor{LightBlue1}\hl{0.5395}&\sethlcolor{LightBlue1}\hl{0.1275}\\
&Moon&\sethlcolor{LightBlue1}\hl{0.6154}&\sethlcolor{LightSkyBlue}\hl{0.1940}&\sethlcolor{LightSkyBlue}\hl{0.4675}&\sethlcolor{LightSkyBlue}\hl{0.1192}&0.5756&0.1401&0.5047&0.1203&0.5392&0.1274\\

\hline
\multirow{5}*{\rotatebox{90}{PSM}}&Original&\sethlcolor{LightSkyBlue}\hl{0.7830}&\sethlcolor{LightSkyBlue}\hl{0.5423}&0.6063&0.4306&0.6313&\sethlcolor{LightBlue1}\hl{0.4715}&0.7389&\sethlcolor{LightSkyBlue}\hl{0.5286}&0.5792&0.3566\\
&FedAvg&0.5826&0.3674&0.6068&0.4308&\sethlcolor{LightBlue1}\hl{0.6693}&\sethlcolor{LightSkyBlue}\hl{0.4794}&\sethlcolor{LightBlue1}\hl{0.7486}&0.4332&\sethlcolor{LightBlue1}\hl{0.5804}&\sethlcolor{LightBlue1}\hl{0.3601}\\
&FedProx&0.5895&0.3728&0.6073&0.4318&0.5682&0.3880&0.7138&\sethlcolor{LightBlue1}\hl{0.4672}&0.4336&0.2408\\
&Scaffold&\sethlcolor{LightBlue1}\hl{0.7606}&\sethlcolor{LightBlue1}\hl{0.5044}&\sethlcolor{LightSkyBlue}\hl{0.7036}&\sethlcolor{LightSkyBlue}\hl{0.4737}&\sethlcolor{LightSkyBlue}\hl{0.6792}&0.4524&\sethlcolor{LightSkyBlue}\hl{0.7494}&0.4357&0.5006&0.2787\\
&Moon&0.5590&0.3697&\sethlcolor{LightBlue1}\hl{0.6111}&\sethlcolor{LightBlue1}\hl{0.4360}&0.6567&0.4237&0.7465&0.4583&\sethlcolor{LightSkyBlue}\hl{0.5833}&\sethlcolor{LightSkyBlue}\hl{0.3636}\\

\Xhline{1pt}

\end{tabular}
}
\end{center}
\label{main_aucroc_aucpr}
\end{table*}

\section{Experiments}

\subsection{Experiment Settings}

For all the datasets, we normalize the data by scaling the value range between 0 and 1 for each dimension respectively. It should be noticed that, for SMD and SMAP, we normalize all the time series uniformly. Inspired by parameter settings in \cite{moon}, we set the proximal weight of FedProx to be 0.01, and set the hyper-parameter "temperature" and the weight of contrastive loss in MOON\cite{moon} to be 0.5 and 1, respectively. For the hyper-parameter settings in time series anomaly detection algorithms, we use the default values as recommended in the original papers if appropriate.

\subsection{Evaluation Metrics}
We choose AUC-ROC (Area under the ROC Curve) and AUC-PR (Area under the Precision-Recall Curve) \cite{zhou2021feature} as evaluation metrics. AUC-ROC is a commonly-used metric for anomaly detection. AUC-PR is the area under the curve representing the correlation between Precision and Recall. 

We also use Precision, Recall and F1-score as evaluation metrics. They are calculated as

\begin{small}
\begin{equation}
    Precision = \frac{TP}{TP+FP}
\end{equation}
\begin{equation}
    Recall = \frac{TP}{TP + FN}
\end{equation}
\end{small}\\

\begin{equation}
    F1 = \frac{2 \times (Precision \times Recall)}{Precision + Recall}
\end{equation}

\noindent where $TP$ is the number of correctly detected anomalous time points, $FP$ is the number of normal time points judged to be anomalous, and $FN$ is the number of time points that are wrongly judged to be normal. Following the literature \cite{audibert2020usad}, we search the best threshold that has the highest F1-score for each experiment.

In the experiments, we also present the adjusted performance metrics, following \cite{audibert2020usad}, \cite{tuli2022tranad}. That is, if one or more anomalous points in an anomalous sub-sequence are judged as abnormal, each anomalous time point in this sub-sequence is considered to be successfully detected.

\subsection{Time series anomaly detection performance with/without federated learning}

To analyze the feasibility of federated learning for typical
time series anomaly detection algorithms, we evaluate the 5 centralized time series anomaly detection methods  and their combinations with 4 different federated learning methods on 3 datasets. 

As shown in Table \ref{main_aucroc_aucpr}, \ref{main_precision_recall} and \ref{main_f1_f1adjusted}, for SMD and SMAP dataset, federated learning clients and time series are in a one-to-one correspondence. Note that, for Table \ref{main_precision_recall} and \ref{main_f1_f1adjusted}, the performance metrics are the adjusted ones as described in Section IV. B. For PSM, time points are assigned consecutively to 24 clients following a Dirichlet distribution with $\beta$ set to 0.5. We surprisingly observe that, in some conditions, federated learning performs better than centralized learning. We hypothesis that federated learning can bring some regularization effect for the training of time series anomaly detection.

\begin{table*}[ht]
\caption{Detailed Results of FedTADBench in terms of Precision and Recall.  We highlight the best result and the second best result  in blue and light blue, respectively. Note that the precision and recall are adjusted metrics as described in Section IV. B.}
\scriptsize
\begin{center}
\newcommand\mtc[1]{\multicolumn{2}{c}{#1}}
\newcommand\mtrtotwo[1]{\multirow{2}*{#1}}

\newcommand\linone{\Xcline{3-12}{0.4pt}}
\resizebox{\textwidth}{!}{
\begin{tabular}{cl|cc cc cc cc cc}
\Xhline{1pt}
\multicolumn{2}{c|}{\mtrtotwo{{Method}}}&\mtc{{DeepSVDD}}&\mtc{{LSTM-AE}}&\mtc{{USAD}}&\mtc{{GDN}}&\mtc{{TranAD}}\\

\makebox[-10pt][c]{}&\makebox[0.03\textwidth][c|]{}&{Precision}&{Recall}&{Precision}&{Recall}&{Precision}&{Recall}&{Precision}&{Recall}&{Precision}&{Recall}\\

\hline
\multirow{5}*{\rotatebox{90}{SMD}}&Original&\sethlcolor{LightBlue1}\hl{0.6921}&0.2754&0.4102&\sethlcolor{LightBlue1}\hl{0.4269}&0.4495&\sethlcolor{LightSkyBlue}\hl{0.4177}&\sethlcolor{LightSkyBlue}\hl{0.6024}&\sethlcolor{LightSkyBlue}\hl{0.7665}&0.5915&0.3637\\
&FedAvg&0.6221&0.4820&\sethlcolor{LightBlue1}\hl{0.5659}&0.2478&\sethlcolor{LightSkyBlue}\hl{0.5911}&\sethlcolor{LightBlue1}\hl{0.4149}&0.3966&\sethlcolor{LightBlue1}\hl{0.7414}&\sethlcolor{LightBlue1}\hl{0.6869}&\sethlcolor{LightBlue1}\hl{0.4920}\\
&FedProx&0.6038&\sethlcolor{LightSkyBlue}\hl{0.8152}&\sethlcolor{LightSkyBlue}\hl{0.5661}&0.2478&0.3441&0.3134&0.3755&0.6013&0.3845&0.3950\\
&Scaffold&0.5245&0.1746&0.3268&\sethlcolor{LightSkyBlue}\hl{0.5708}&0.4820&0.4113&0.2966&0.6166&\sethlcolor{LightSkyBlue}\hl{0.7423}&0.3620\\
&Moon&\sethlcolor{LightSkyBlue}\hl{0.8567}&\sethlcolor{LightBlue1}\hl{0.8060}&0.5485&0.2674&\sethlcolor{LightBlue1}\hl{0.5431}&0.3128&\sethlcolor{LightBlue1}\hl{0.4563}&0.5745&0.6624&\sethlcolor{LightSkyBlue}\hl{0.4926}\\

\hline
\multirow{5}*{\rotatebox{90}{SMAP}}&Original&0.7118&0.8625&0.6900&\sethlcolor{LightSkyBlue}\hl{0.4311}&0.9549&0.5631&0.9538&\sethlcolor{LightBlue1}\hl{0.5613}&0.8457&\sethlcolor{LightSkyBlue}\hl{0.8096}\\
&FedAvg&0.7947&0.7989&0.6902&\sethlcolor{LightSkyBlue}\hl{0.4311}&\sethlcolor{LightBlue1}\hl{0.9574}&0.5631&\sethlcolor{LightSkyBlue}\hl{0.9796}&0.5508&\sethlcolor{LightBlue1}\hl{0.9018}&0.5644\\
&FedProx&\sethlcolor{LightSkyBlue}\hl{0.9254}&0.5572&\sethlcolor{LightBlue1}\hl{0.6907}&\sethlcolor{LightSkyBlue}\hl{0.4311}&0.3078&\sethlcolor{LightSkyBlue}\hl{0.9159}&\sethlcolor{LightBlue1}\hl{0.9784}&0.5529&0.7813&\sethlcolor{LightBlue1}\hl{0.5717}\\
&Scaffold&0.7774&\sethlcolor{LightSkyBlue}\hl{0.8848}&0.6902&\sethlcolor{LightSkyBlue}\hl{0.4311}&\sethlcolor{LightSkyBlue}\hl{0.9578}&0.5616&0.9521&0.5539&\sethlcolor{LightSkyBlue}\hl{0.9029}&0.5694\\
&Moon&\sethlcolor{LightBlue1}\hl{0.8998}&\sethlcolor{LightBlue1}\hl{0.8811}&\sethlcolor{LightSkyBlue}\hl{0.8160}&0.3905&0.9555&\sethlcolor{LightBlue1}\hl{0.5640}&0.9531&\sethlcolor{LightSkyBlue}\hl{0.5635}&\sethlcolor{LightBlue1}\hl{0.9018}&0.5644\\

\hline
\multirow{5}*{\rotatebox{90}{PSM}}&Original&0.8264&\sethlcolor{LightBlue1}\hl{0.8819}&\sethlcolor{LightBlue1}\hl{0.6997}&0.6922&0.5644&\sethlcolor{LightSkyBlue}\hl{0.9269}&\sethlcolor{LightBlue1}\hl{0.8240}&\sethlcolor{LightBlue1}\hl{0.9107}&0.7605&\sethlcolor{LightBlue1}\hl{0.8543}\\
&FedAvg&\sethlcolor{LightBlue1}\hl{0.8749}&0.8198&0.6795&0.6922&\sethlcolor{LightBlue1}\hl{0.6720}&0.8989&0.6781&\sethlcolor{LightSkyBlue}\hl{0.9590}&\sethlcolor{LightSkyBlue}\hl{0.8214}&0.8381\\
&FedProx&0.8476&0.8730&0.6796&0.6922&0.4758&\sethlcolor{LightBlue1}\hl{0.9192}&\sethlcolor{LightSkyBlue}\hl{0.9332}&0.9000&0.6360&0.6215\\
&Scaffold&\sethlcolor{LightSkyBlue}\hl{0.9717}&0.7128&\sethlcolor{LightSkyBlue}\hl{0.7676}&\sethlcolor{LightSkyBlue}\hl{0.7562}&\sethlcolor{LightSkyBlue}\hl{0.6727}&0.9003&0.6684&0.8852&0.2776&\sethlcolor{LightSkyBlue}\hl{1.0000}\\
&Moon&0.5789&\sethlcolor{LightSkyBlue}\hl{0.9347}&0.6583&\sethlcolor{LightBlue1}\hl{0.6923}&0.6335&0.9023&0.7890&0.8920&\sethlcolor{LightBlue1}\hl{0.8005}&0.8378\\

\Xhline{1pt}

\end{tabular}
}
\end{center}
\label{main_precision_recall}
\end{table*}

\begin{figure}[htbp]
\centerline{\includegraphics[width=1.0\linewidth]{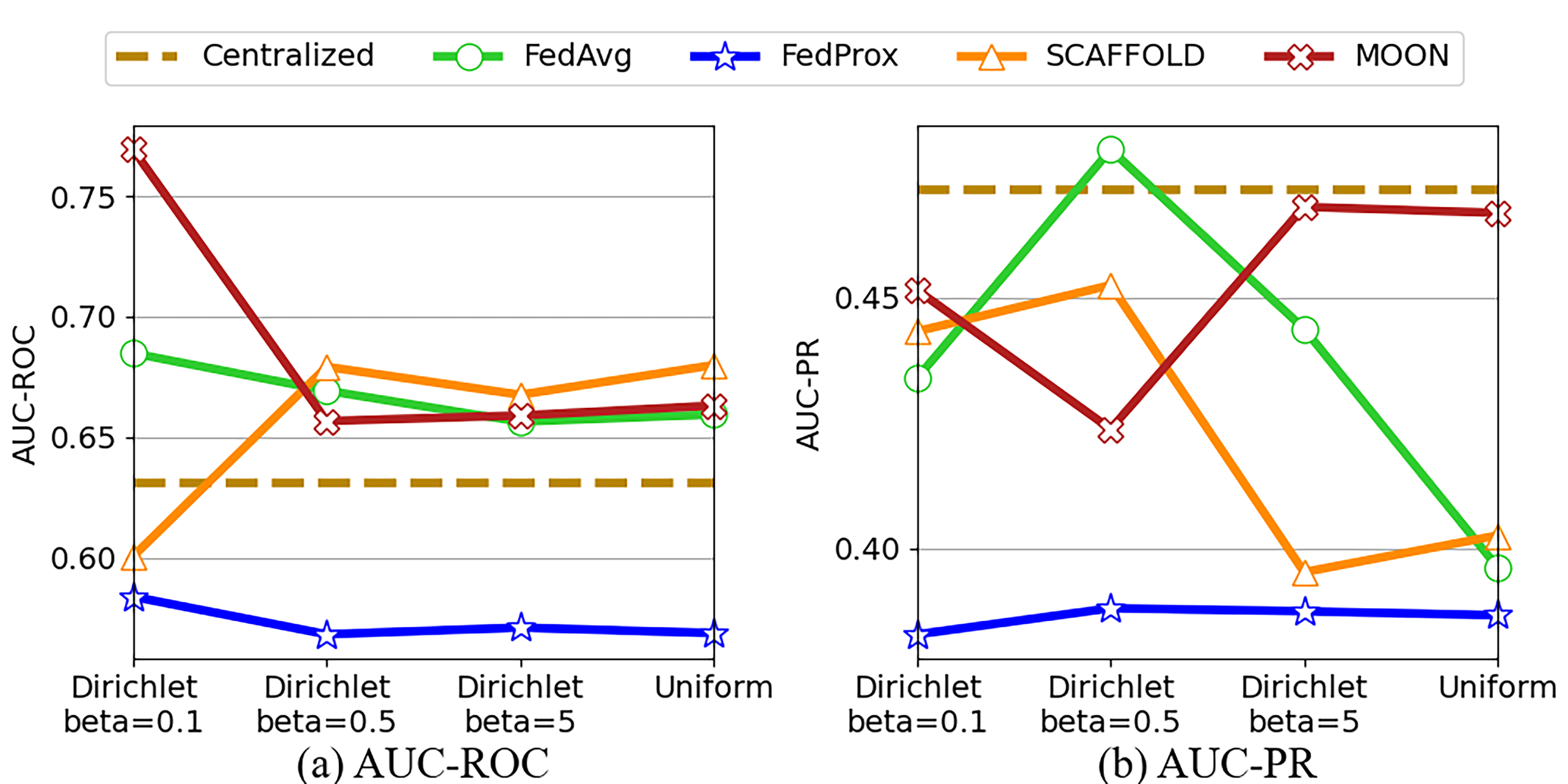}}
\caption{AUC-ROC and AUC-PR of USAD on PSM with different partitions of data in clients.}
\label{fig_beta_auc}
\end{figure}

\begin{figure}[htbp]
\centerline{\includegraphics[width=1.0\linewidth]{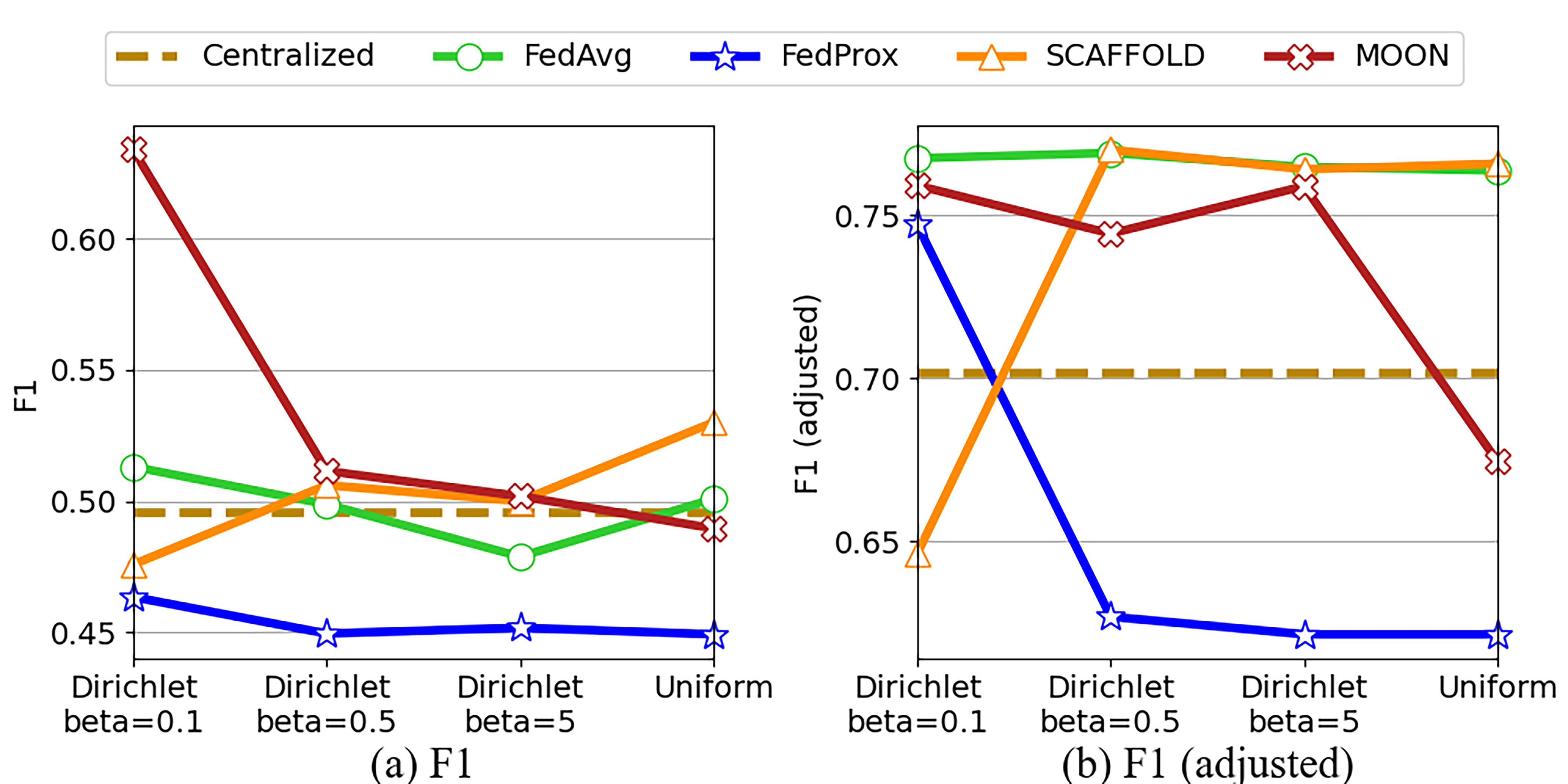}}
\caption{F1 and F1 (adjusted) of USAD on PSM with different partitions of data in clients.}
\label{fig_beta_f1}
\end{figure}

\subsection{Time series anomaly detection performance with federated learning when data being differently partitioned in clients}

To explore  the
influence of the heterogeneity between clients to federated
time series anomaly detection, we evaluate federated time series anomaly detection with different partitions of data in clients.

Specifically, we employ 4 different partitions in the client-generating procedure. First of all, we employ an average distribution. The numbers of timestamps in different clients are equal or extremely close. In addition, Dirichlet distributions with 3 different $\beta$ values (0.1, 0.5 and 5) are employed, which is the same as the settings in \cite{moon}. We perform experiments on federated learning method FedAvg as an example. The AUC-ROC and AUC-PR results are shown in Fig. \ref{fig_beta_auc}, and the F1 and F1 (adjusted) scores are shown in Fig. \ref{fig_beta_f1}.

It could be observed that the performance of federated learning frameworks  is robust to the change of  $\beta$, i.e., the unbalanced data distribution does not affect the performance of time series anomaly detection much. In federated learning, the number of data essentially affects the updates of the model in one global epoch, and this result implies that,  for time series anomaly detection, the number of updates does not cause much heterogeneity between local models.

\subsection{Compatibility between different federated learning frameworks and time series}

We are also interested in the compatibility between popular federated learning methods and typical time series anomaly detection algorithms.

As shown in Table \ref{main_aucroc_aucpr}, \ref{main_precision_recall} and \ref{main_f1_f1adjusted}, we can see that the performance of FedProx varies significantly under different conditions. FedProx builds a strong regularization on the change of local model parameters. We argue this may not be optimal because that, in some conditions, we may require certain parts of the model parameters to adapt to the data change efficiently. For example, USAD is a two-stage algorithm, we may need the parameters of stage one to change properly. However, FedProx limits the change of parameters in both stages, which may make FedProx to be the worst federated learning framework for USAD. On the other hand, MOON constrains the change of model in a more relaxed manner by constraining the change of features.  Moreover, the performance of FedAvg and SCAFFOLD is mostly in the middle level because they introduce fewer fixed assumptions.

\begin{figure}[htbp]
\centerline{\includegraphics[width=1.0\linewidth]{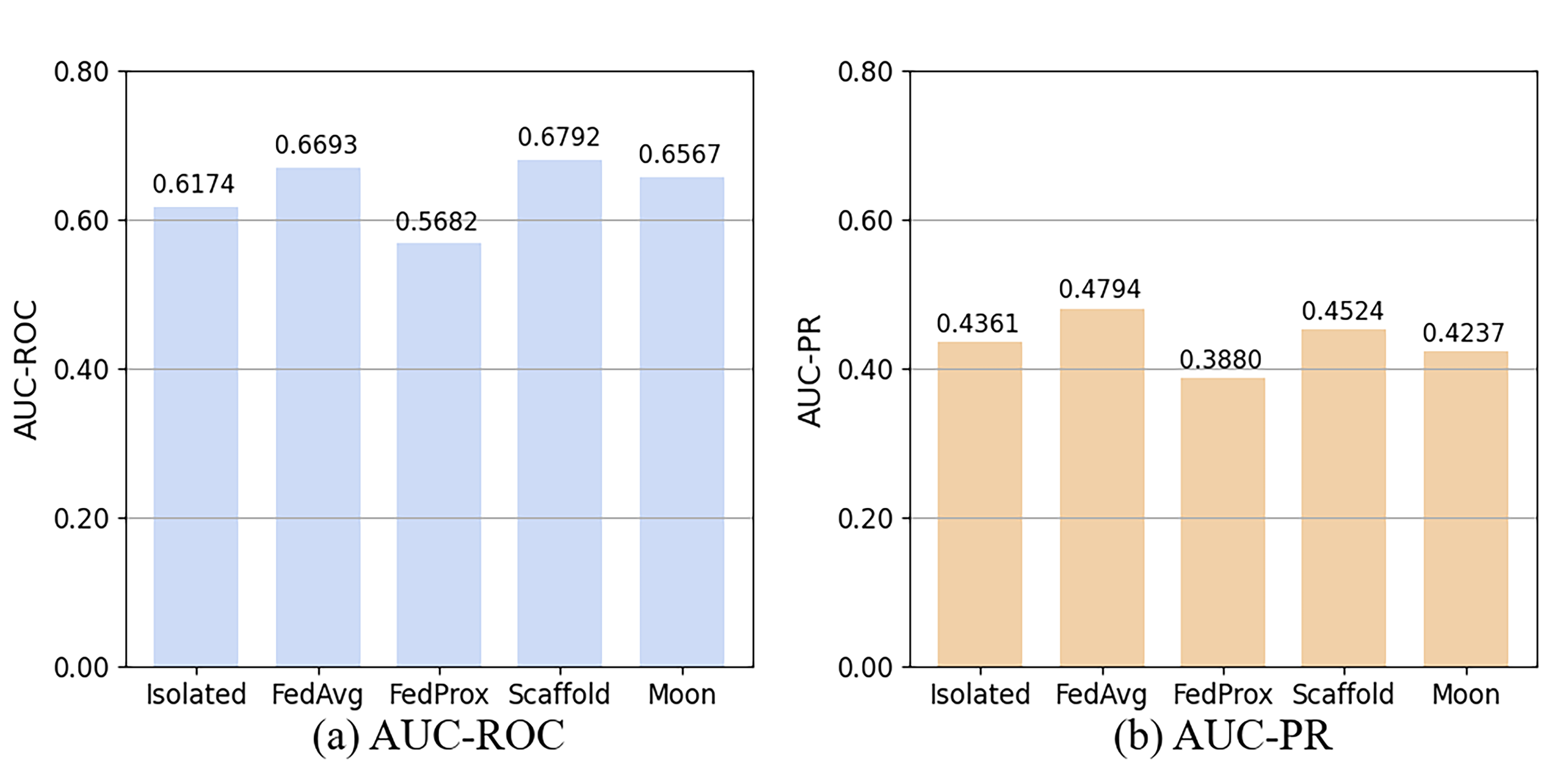}}
\caption{AUC-ROC and AUC-PR of isolated and federated training for USAD on PSM.}
\label{fig_average_auc}
\end{figure}

\begin{figure}[htbp]
\centerline{\includegraphics[width=1.0\linewidth]{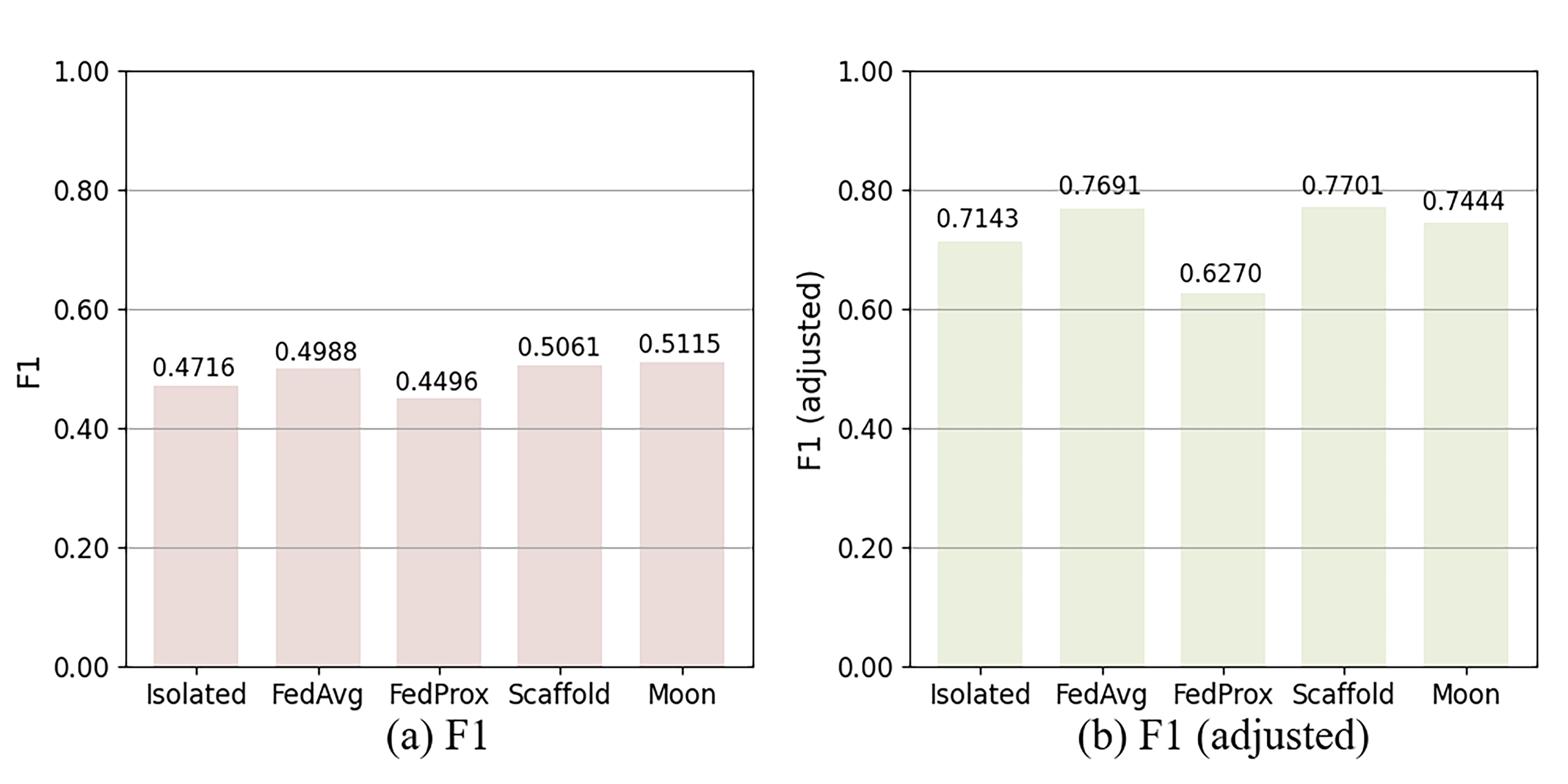}}
\caption{F1 and F1 (adjusted) of isolated and federated training for USAD on PSM.}
\label{fig_average_f1}
\end{figure}

\begin{table*}[htbp]
\caption{ Detailed Results of FedTADBench in terms of F1 and F1 (adjusted).  We highlight the best result and the second best result  in blue and light blue, respectively.}
\begin{center}
\newcommand\mtc[1]{\multicolumn{2}{c}{#1}} 
\newcommand\mtrtotwo[1]{\multirow{2}*{#1}} 
\newcommand\linone{\Xcline{3-12}{0.4pt}}

\resizebox{\linewidth}{!}{
\begin{tabular}{cl|cc cc cc cc cc}
\Xhline{1pt}
\multicolumn{2}{c|}{\mtrtotwo{{Method}}}&\mtc{{DeepSVDD}}&\mtc{{LSTM-AE}}&\mtc{{USAD}}&\mtc{{GDN}}&\mtc{{TranAD}}\\

\multicolumn{2}{c|}{}&{F1}&{F1(adjusted)}&{F1}&{F1(adjusted)}&{F1}&{F1(adjusted)}&{F1}&{F1(adjusted)}&{F1}&{F1(adjusted)}\\

\hline
\multirow{5}*{\rotatebox{90}{SMD}}&Original&0.1290&0.3941&0.1373&\sethlcolor{LightSkyBlue}\hl{0.4184}&\sethlcolor{LightSkyBlue}\hl{0.1872}&0.4330&\sethlcolor{LightSkyBlue}\hl{0.1777}&\sethlcolor{LightSkyBlue}\hl{0.6747}&0.1435&0.4504\\
&FedAvg&\sethlcolor{LightBlue1}\hl{0.1792}&0.5431&0.1413&0.3446&\sethlcolor{LightBlue1}\hl{0.1813}&\sethlcolor{LightSkyBlue}\hl{0.4876}&\sethlcolor{LightBlue1}\hl{0.1490}&\sethlcolor{LightBlue1}\hl{0.5168}&\sethlcolor{LightBlue1}\hl{0.1829}&\sethlcolor{LightSkyBlue}\hl{0.5734}\\
&FedProx&0.1628&\sethlcolor{LightBlue1}\hl{0.6937}&\sethlcolor{LightBlue1}\hl{0.1424}&0.3447&0.1095&0.3280&0.1430&0.4623&0.1100&0.3896\\
&Scaffold&0.1725&0.2620&\sethlcolor{LightSkyBlue}\hl{0.1518}&\sethlcolor{LightBlue1}\hl{0.4156}&0.1717&\sethlcolor{LightBlue1}\hl{0.4439}&0.1346&0.4005&0.1490&0.4867\\
&Moon&\sethlcolor{LightSkyBlue}\hl{0.2188}&\sethlcolor{LightSkyBlue}\hl{0.8306}&0.1408&0.3595&0.1793&0.3970&0.1316&0.5086&\sethlcolor{LightSkyBlue}\hl{0.1833}&\sethlcolor{LightBlue1}\hl{0.5650}\\

\hline
\multirow{5}*{\rotatebox{90}{SMAP}}&Original&0.2804&0.7799&0.2342&\sethlcolor{LightBlue1}\hl{0.5307}&0.2805&0.7084&0.2584&\sethlcolor{LightBlue1}\hl{0.7067}&\sethlcolor{LightSkyBlue}\hl{0.2917}&\sethlcolor{LightSkyBlue}\hl{0.8272}\\
&FedAvg&\sethlcolor{LightSkyBlue}\hl{0.3303}&0.7968&0.2272&\sethlcolor{LightBlue1}\hl{0.5307}&\sethlcolor{LightSkyBlue}\hl{0.2851}&\sethlcolor{LightBlue1}\hl{0.7091}&0.2665&0.7050&0.2782&0.6943\\
&FedProx&0.2323&0.6956&\sethlcolor{LightBlue1}\hl{0.2445}&\sethlcolor{LightSkyBlue}\hl{0.5309}&0.2332&0.4608&\sethlcolor{LightBlue1}\hl{0.2696}&0.7065&0.2744&0.6602\\
&Scaffold&\sethlcolor{LightBlue1}\hl{0.2922}&\sethlcolor{LightBlue1}\hl{0.8276}&0.2272&0.4311&\sethlcolor{LightBlue1}\hl{0.2825}&0.7081&\sethlcolor{LightSkyBlue}\hl{0.2892}&0.7003&\sethlcolor{LightBlue1}\hl{0.2784}&\sethlcolor{LightBlue1}\hl{0.6984}\\
&Moon&0.2840&\sethlcolor{LightSkyBlue}\hl{0.8904}&\sethlcolor{LightSkyBlue}\hl{0.2550}&0.5282&0.2795&\sethlcolor{LightSkyBlue}\hl{0.7093}&0.2671&\sethlcolor{LightSkyBlue}\hl{0.7082}&0.2782&0.6943\\

\hline
\multirow{5}*{\rotatebox{90}{PSM}}&Original&\sethlcolor{LightSkyBlue}\hl{0.6260}&\sethlcolor{LightBlue1}\hl{0.8532}&0.4414&\sethlcolor{LightBlue1}\hl{0.6959}&0.4957&0.7015&\sethlcolor{LightSkyBlue}\hl{0.5924}&\sethlcolor{LightBlue1}\hl{0.8652}&0.4451&0.8047\\
&FedAvg&0.4356&0.8464&0.4415&0.6858&0.4988&\sethlcolor{LightBlue1}\hl{0.7691}&0.5711&0.7945&\sethlcolor{LightBlue1}\hl{0.4472}&\sethlcolor{LightSkyBlue}\hl{0.8296}\\
&FedProx&0.4617&\sethlcolor{LightSkyBlue}\hl{0.8601}&0.4420&0.6858&0.4496&0.6270&0.5336&\sethlcolor{LightSkyBlue}\hl{0.9163}&0.4345&0.6287\\
&Scaffold&\sethlcolor{LightBlue1}\hl{0.6178}&0.8224&\sethlcolor{LightSkyBlue}\hl{0.5500}&\sethlcolor{LightSkyBlue}\hl{0.7619}&\sethlcolor{LightBlue1}\hl{0.5061}&\sethlcolor{LightSkyBlue}\hl{0.7701}&\sethlcolor{LightBlue1}\hl{0.5851}&0.7616&0.4345&0.4345\\
&Moon&0.4345&0.7150&\sethlcolor{LightBlue1}\hl{0.4501}&0.6749&\sethlcolor{LightSkyBlue}\hl{0.5115}&0.7444&0.5564&0.8374&\sethlcolor{LightSkyBlue}\hl{0.4483}&\sethlcolor{LightBlue1}\hl{0.8187}\\

\Xhline{1pt}

\end{tabular}
}
\end{center}
\label{main_f1_f1adjusted}
\end{table*}

\subsection{Time series anomaly detection performance under isolated and federated settings}

To demonstrate whether federated learning is meaningful, we compare the time series anomaly detection performance under federated learning and isolated training, respectively.

As shown in Fig. \ref{fig_average_auc} and Fig. \ref{fig_average_f1}, we evaluate the performance of USAD on PSM under isolated and federated settings. It is obvious that in most instances, the performance with federated learning is better than that when training in isolation, which demonstrates the effectiveness of federated learning.

\subsection{Training time without/with federated learning}

To better guide the choice of federated learning methods for time series anomaly detection, we also evaluate the training time of typical time series anomaly detection algorithm with different federated learning strategies.

Training time of one global epoch of USAD on PSM in different federated learning strategies is shown in Table \ref{runtime_table}. For this evaluation, the number of local epochs in each client for all federated learning methods is 10. All the experiments are conducted on NVIDIA Geforce RTX 3090 (24GB) GPU. Note that clients in federated and isolated settings are trained in parallel. It could be observed that the training time of FedAvg, whose performance is stable according to the aforementioned analysis, is shorter than that of the other three federated learning methods. MOON takes the longest time for training, due to the complex calculations involved by contrastive learning.

\begin{table}[!h]

\caption{Training time of USAD on PSM.}
\begin{center}
\newcommand\mtc[1]{\multicolumn{2}{c}{#1}}
\newcommand\mtrtotwo[1]{\multirow{2}*{#1}}
\newcommand\linone{\Xcline{3-12}{0.4pt}}
\newcommand\myfont[1]{\fontsize{3.0pt}{\baselineskip}\selectfont{#1}}
\renewcommand\arraystretch{0.4}

\resizebox{0.95\columnwidth}{!}{

\begin{tabular}{ccc}
\Xhline{0.35pt}
\myfont{Learing}&\myfont{Federated}&\myfont{Training Time/s}\\
\myfont{Manner}&\myfont{Learning}&\myfont{(one global epoch)}\\
\specialrule{0.1pt}{0pt}{0pt}
\myfont{Centralized}&\myfont{N}&\myfont{20.99}\\
\myfont{Isolated}&\myfont{N}&\myfont{2.64}\\
\myfont{FedAvg}&\myfont{Y}&\myfont{27.51}\\
\myfont{FedProx}&\myfont{Y}&\myfont{37.29}\\
\myfont{Scaffold}&\myfont{Y}&\myfont{38.60}\\
\myfont{Moon}&\myfont{Y}&\myfont{64.88}\\
\specialrule{0.35pt}{0pt}{1pt}

\end{tabular}
}
\end{center}
\label{runtime_table}
\end{table}

\section{Conclusion}
In this paper, we present the first benchmark for time series anomaly detection under federated learning frameworks. Our benchmark covers 5 time series anomaly detection algorithms, 4 federated learning frameworks, and 3 time series anomaly detection datasets. Based on our experiments, we analyze the effects federated learning brings in, the influence of unbalanced data distribution and the compatibility between different federated learning frameworks and time series. These analyses can provide some guiding information so that researchers can better choose time series anomaly detection algorithms and federated learning frameworks when performing time series anomaly detection under federated learning frameworks. Our work may help develop a federated learning framework that takes the characteristic of time series anomaly detection into account.

\section{Acknowledgment}

This work was supported in part by National Natural Science Foundation of China (NSFC) with grant number 62171302 and U19A2052.

\bibliographystyle{IEEEtran}
\bibliography{FedTADBench_arxiv.bbl}{}

\end{document}